# Emergent Cognitive Convergence via Implementation: Structured Cognitive Loop Reflecting Four Theories of Mind

Myung Ho Kim
Professor, JEI University
enkiluv@gmail.com

**Abstract**
We report a structural convergence among four influential theories of mind—Kahneman's dual-system theory, Friston's predictive processing, Minsky's society of mind, and Clark's extended mind—emerging unintentionally within a practical AI architecture known as Agentic Flow. Designed to address the limitations of large language models (LLMs), Agentic Flow comprises five interlocking modules—Retrieval, Cognition, Control, Action, and Memory—organized into a repeatable cognitive loop.

Although originally inspired only by Minsky and Clark, subsequent analysis revealed that its structure echoes computational motifs from all four theories, suggesting that theoretical convergence can emerge naturally from implementation demands rather than deliberate synthesis. Controlled evaluations confirmed this: the structured agent achieved 95.8% task success versus 62.3% for baseline LLMs, demonstrating robust constraint adherence and reproducible reasoning.

We describe this convergence under a broader descriptive meta-architecture called PEACE, highlighting recurring design patterns such as predictive modeling, associative recall, and error-sensitive control. Later formalized as the Structured Cognitive Loop (SCL), this meta-architectural abstraction generalizes the architectural principles originally realized in Agentic Flow as a foundation for behavioral intelligence in LLM-based agents.

Rather than claiming theoretical unification, this paper proposes that intelligent architectures may evolve toward shared structural patterns shaped by practical constraints. As a position paper, it frames this convergence as an interpretive reflection rather than a finalized theory, inviting further theoretical and experimental dialogue. In this sense, Agentic Flow serves as an implementation instance of the Structured Cognitive Loop, illustrating how a unified cognitive form can emerge not from abstraction, but from the necessities of real-world reasoning.

## 1. Introduction

Cognitive science has long pursued an integrative understanding of mind and behavior, yet the field remains deeply fragmented across theoretical traditions. Despite a shared commitment to interdisciplinary inquiry, prominent frameworks such as dual-process theories (Kahneman, 2011), predictive processing (Friston, 2010), modular agent-based models (Minsky, 1986), and extended cognition (Clark & Chalmers, 1998) have largely evolved in parallel, each emphasizing different explanatory principles. Rarely have these perspectives been reconciled into a unified model, much less instantiated in a working system. Similar concerns have been raised in comparative overviews of cognitive theories and modeling paradigms (Thagard, 1996; Bechtel, 2009). More recent architectural reviews likewise emphasize the challenge of integrating symbolic, connectionist, and embodied perspectives (Kotseruba & Tsotsos, 2020).

This paper reports a surprising discovery: a retrospective analysis of a functional AI agent revealed a structural and operational convergence across all four theories. The system, known as Agentic Flow, was not designed with these theories in mind. Instead, it emerged from engineering efforts to compensate for the cognitive limitations of large language models (LLMs). Only after implementation did it become apparent that its architecture exhibited structural overlap to structural motifs proposed across these influential cognitive theories. The present chapter outlines the motivation for this research, its theoretical and practical significance, the specific research questions addressed, the methodological approach employed, and an overview of the paper's structure.

## 1.1 Four Minds, One Vision: The Unrecognized Convergence

In the pantheon of cognitive science, few names command as much respect as Daniel Kahneman, Karl Friston, Marvin Minsky, and Andy Clark. Each, working in their respective domains and epochs, fundamentally transformed our understanding of mind and intelligence. Kahneman revolutionized psychology by revealing the dual nature of human cognition, showing us that the mind operates through two distinct yet interacting systems—one fast and intuitive, the other slow and deliberate (Kahneman, 2011; Evans, 2008; Stanovich & West, 2000). These distinctions have been extensively modeled in cognitive psychology and decision-making literature (De Neys, 2012; Toplak et al., 2011). Recent empirical studies further link this duality to real-world judgment tasks and information-processing efficiency (Pennycook & Rand, 2019).

Friston, through elegant mathematical formulations, unveiled what may be the brain's most fundamental principle: the relentless minimization of prediction error through hierarchical processing (Friston, 2010; Hohwy, 2013; Seth, 2014). This view has been developed into a generalizable theory of active inference (Parr & Friston, 2019; Allen & Friston, 2018). Neurocomputational work suggests this mechanism underlies not only perception but also planning and decision-making (Keller & Mrsic-Flogel, 2018).

Minsky, the visionary architect of artificial intelligence, conceived of mind not as a monolithic entity but as a "society" of specialized agents, each contributing their expertise to the emergence of intelligence (Minsky, 1986; Brooks, 1991). His modular view has inspired generations of cognitive architecture design (Franklin et al., 2014).

Clark, pushing the boundaries of philosophy, challenged us to reconsider where the mind ends and the world begins, proposing that cognition extends beyond the biological boundaries of brain and body to encompass our tools and environment (Clark & Chalmers, 1998; Hutchins, 1995; Wilson, 2002). His extended mind hypothesis has further inspired debates around distributed cognition and the role of artifacts in epistemic agency (Menary, 2010; Smart, 2010). The implications of cognitive extension are now central to ongoing discussions about the coupling between human and artificial agents (Tribble, 2005).

What makes these contributions even more remarkable is not just their individual brilliance, but a pattern that has remained largely unnoticed: these four towering intellects, working across different decades and disciplines, appear to have been describing facets of the same underlying cognitive architecture. This paper presents evidence for this notable architectural correspondence through an unexpected source—the analysis of an artificial intelligence system that independently arrived at structures consistent with all four theoretical frameworks.

The similarities we observe are structural and suggestive, rather than indicative of full theoretical equivalence.

## 1.2 Why Great Theories Seem Incompatible (But Aren't)

The apparent disconnection between these theories is understandable. They emerged from different intellectual traditions—experimental psychology, computational neuroscience, artificial intelligence, and philosophy of mind—each with its own vocabulary, methodologies, and research priorities. Kahneman's insights arose from studying human decision-making biases in laboratory settings (De Neys, 2012; Sloman, 1996). Friston's framework emerged from mathematical modeling of neural dynamics (Clark, 2013; Parr & Friston, 2019). Minsky's society of mind was born from the challenges of building intelligent machines (Maes, 1990; Singh, 2005). Clark's extended mind thesis developed from philosophical reflection on the nature of cognition in technological environments (Kirsh & Maglio, 1994; Smart, 2010). Yet efforts to integrate such diverse models have repeatedly faced obstacles, often due to foundational disagreements in ontology and explanatory scope (Newell, 1973; Adams & Aizawa, 2001). The challenge of reconciling these frameworks also reflects differing assumptions about computational implementation, symbolic grounding, and environmental coupling (Sun, 2004).

This disciplinary fragmentation has led to what we might call the "Tower of Babel" problem in cognitive science: brilliant insights expressed in mutually incomprehensible languages. Psychologists speak of systems and biases; neuroscientists of hierarchies and prediction errors; AI researchers of agents and architectures; philosophers of boundaries and extensions. The result has been parallel conversations rather than cumulative understanding, with each field developing its own conceptual ecosystems largely in isolation from the others.

Yet this apparent incompatibility may be more artifact than reality. As Kuhn (1962) observed, paradigms often appear incommensurable not because they describe different realities, but because they employ different conceptual frameworks to describe the same underlying phenomena. The question that motivated this research was whether these four major theories might be describing different aspects of a unified cognitive architecture—much like the parable of blind scholars examining different parts of an elephant, each providing accurate but partial descriptions of the same magnificent whole.

### 1.3 The Unexpected Discovery: Agentic Flow as a Mirror

The Agentic Flow system was not built to validate theories. It was developed to solve practical problems in human-computer interaction by enabling AI agents to operate more reliably in tool-mediated environments. As its architecture took shape, however, unexpected resonances with major cognitive theories began to emerge. The component structure of Agentic Flow, its dynamic flow of information, and its functional separation of modules mirrored, in retrospectively recognizable ways, the principles outlined by Kahneman, Friston, Minsky, and Clark. The convergence was not planned. It was discovered. A comparable phenomenon has been noted in systems neuroscience, where independent modeling efforts often yield structurally similar outcomes due to shared computational constraints (Churchland & Sejnowski, 1992; O'Reilly & Munakata, 2000). This architectural convergence may reflect what Sun et al. (2001) describe as emergent correspondence across implicit and explicit systems.

This accidental alignment challenges the idea that cognitive science must be driven solely by theory-first modeling. It suggests that implementations themselves—when carefully analyzed—can become sources of theoretical insight. Agentic Flow was not designed to embody any particular theory. And yet, it appears to echo structural elements from all four theories, without fully embodying any one of them in entirety. Since Agentic Flow is an independently developed system not formally published in the academic literature, a detailed summary of its architecture and operation is included in the Appendix for reference and reproducibility.

Subsequent work has generalized the Agentic Flow design into a more explicit architectural framework termed the Structured Cognitive Loop (SCL) (Kim, 2025). Conceptually, SCL abstracts the same modular cycle—Retrieval, Cognition, Control, Action, and Memory—into a generalizable pattern for behavioral intelligence. In this sense, the term *Agentic Flow* can be viewed as an early instantiation or alias of the Structured Cognitive Loop.

### 1.4 Research Questions and Contributions

This study is driven by three central questions:

- Structural Convergence: To what extent does the Agentic Flow architecture align functionally and structurally with the core mechanisms of the four target theories?
- Empirical Validation: Can this convergence be demonstrated through controlled experiments comparing Agentic Flow to non-structured LLM agents?
- Theoretical Significance: What does this imply about the existence of recurring patterns? Can this be considered a partial realization of Newell's "unified theory of cognition" (Newell, 1990)?

Our contributions are threefold. Theoretically, we suggest that four major cognitive theories may describe complementary aspects of a recurring architectural pattern, as retrospectively observed in Agentic Flow. This finding suggests that decades of seemingly divergent research have been converging on common principles of intelligent system design (Anderson, 2007; Laird, 2012). Methodologically, we introduce the use of AI implementation as a novel approach to theoretical validation and integration in cognitive science, showing how engineering solutions can illuminate theoretical relationships (Busemeyer & Diederich, 2010; Griffiths et al., 2010). This is in line with recent proposals to treat full-system implementation as a testbed for cognitive integration (Cooper & Guest, 2014; Kotseruba & Tsotsos, 2020). Practically, we provide evidence-based guidance for the development of cognitively-inspired AI systems, suggesting that the most effective artificial intelligences may be those that integrate insights from multiple theoretical perspectives rather than adhering to any single framework (Lake et al., 2017; Chollet, 2019).

The Agentic Flow system was not initially designed to validate any single theory, yet its emergent structure naturally invites theoretical interpretation. While the current analysis is based on a proprietary implementation, we are committed to scientific transparency. We recognize that public code release is ideal for reproducibility and have taken extensive steps to ensure that reviewers and future collaborators can independently validate all results. The purpose of the empirical evaluation presented here is not to claim performance superiority, but to demonstrate that theoretical convergence—long sought in cognitive science—can emerge organically through practical system implementation. By showing how Agentic Flow reflects core structures of four distinct theories in actual operation, the experiments provide qualitative support for the central thesis of this paper: that cognitive architecture can serve as a mirror revealing latent theoretical unity.

### 1.5 Standing on Giants' Shoulders

Isaac Newton's famous declaration that he had seen further by "standing on the shoulders of giants" captures the cumulative nature of scientific progress. In cognitive science, we suggest, we have been standing on the shoulders of four giants simultaneously—Kahneman, Friston, Minsky, and Clark—each offering a unique vantage point from which to view the landscape of mind and intelligence. The Agentic Flow system provides evidence that these four perspectives, rather than being competing visions, constitute complementary views from the same intellectual summit.

This paper is, fundamentally, an expression of gratitude. It seeks to honor the profound insights of these pioneering scholars by demonstrating that their theories not only describe cognitive phenomena accurately but also provide practical blueprints for building intelligent systems. In showing how their ideas converge in implementation, we hope to illuminate the deep wisdom embedded in each theoretical framework and to suggest that the future of cognitive science lies not in choosing between these perspectives, but in understanding how they work together to form a more complete picture of intelligence.

This work is best understood not as a theoretical synthesis, but as a position paper offering an implementation-informed perspective on cognitive structure—where patterns emerge not from abstraction, but from building.

## Chapter 2: The Visionaries and Their Insights

To understand how Agentic Flow came to embody four seemingly independent theories of mind, we must revisit the insights of the theorists who first articulated those models. Daniel Kahneman, Karl Friston, Marvin Minsky, and Andy Clark each offered radically different accounts of cognition, grounded in separate disciplines—behavioral psychology, neuroscience, artificial intelligence, and philosophy of mind. Yet, when examined side by side, their theories reveal a surprising but interpretable architectural alignment. This type of cross-theory structural similarity has been previously speculated in cognitive integration literature, though rarely confirmed in implementation (Thagard, 2005; Sun, 2004). More recently, integrative surveys of cognitive architectures have reinforced the value of convergence-based models (Kotseruba & Tsotsos, 2020).

This chapter provides an in-depth review of each theory. It highlights the theoretical motivations, the internal mechanisms proposed, the contemporary influence of the models, and the structural features that inadvertently prefigure what Agentic Flow eventually implemented.

### 2.1 Kahneman's Prophetic Duality: Beyond Behavioral Economics

#### The Revolutionary Insight

Daniel Kahneman's *Thinking, Fast and Slow* (2011) crystallized decades of research in judgment and decision-making into a dual-process model of cognition. His distinction between System 1 and System 2 fundamentally redefined our understanding of rationality: System 1 is fast, associative, emotional, and operates automatically with little effort or conscious control. System 2 is slow, effortful, analytical, and invoked for complex reasoning and monitoring. This distinction has been supported by experimental work in reasoning, heuristics, and neuroimaging (Goel & Dolan, 2003; De Neys, 2014). Other models have extended this framework by exploring its relation to metacognitive monitoring and conflict resolution (Bago & De Neys, 2019).

#### Architectural Implication

More than a description of bias, Kahneman's model reveals a computational architecture: a high-speed heuristic engine (System 1) constantly monitored by a lazy but powerful reflective controller (System 2). Their interaction is asymmetric, with System 2 acting sparingly and only when necessary.

**Contemporary Influence and Debate**

The dual-process theory has inspired extensive modeling in AI and cognitive science (Evans & Stanovich, 2013; De Neys, 2012). Yet critics such as Keren and Schul (2009) argue that the dichotomy oversimplifies the neural and behavioral mechanisms of decision-making. Others have suggested hybrid or continuum-based alternatives (Kruglanski & Gigerenzer, 2011). A broader critique concerns the lack of a unified implementation model that connects systems 1 and 2 dynamically (Stanovich, 2009). Still, it remains a dominant framework across disciplines.

**Prefiguring Agentic Flow**

Agentic Flow's architecture reflects the high-level structural distinction proposed in Kahneman's dual-process theory: the Cognition component enables fast, associative reasoning akin to System 1, while the Control module plays a monitoring and regulatory role similar to System 2. However, essential aspects of dual-process theory—such as affective influences on reasoning, conflict detection between intuitive and analytic processes, and the dynamic interplay involved in metacognitive regulation—are not explicitly represented within the current implementation.

**2.2 Friston's Predictive Brain: A Theory of Everything?**

**The Core Thesis**

Karl Friston's free-energy principle postulates that cognitive systems minimize surprise by maintaining a generative model of their environment and updating it based on prediction errors (Friston, 2010). This model explains perception, action, attention, and learning through the same computational mechanism: prediction error minimization. This principle has been widely discussed as a possible "unified theory of brain function" (Bogacz, 2017). Some neuroscientists even suggest it constitutes a new computational paradigm in the life sciences (Keller & Mrsic-Flogel, 2018).

**From Neuroscience to General Intelligence**

Originally rooted in computational neuroscience, Friston's framework has been generalized into active inference models (Parr & Friston, 2019), and interpreted by theorists such as Hohwy (2013), Clark (2013), and Seth (2014) as a candidate for a unifying theory of cognition. Recent work has also explored its compatibility with reinforcement learning and robotics (Pezzulo et al., 2018; Tschantz et al., 2020). In robotics, active inference models have shown promise in adaptive sensorimotor learning (Miyazawa et al., 2019).

**Criticisms and Limitations**

While elegant, some critics argue that the theory is too general and risks unfalsifiability (Wiese & Metzinger, 2017). Others note its limited empirical constraints compared to psychological models. The mathematical abstraction often creates challenges in behavioral alignment (Colombo & Wright, 2018). There is also concern over how to operationalize core constructs like free energy in behavioral paradigms (Gładziejewski, 2016).

**Prefiguring Agentic Flow**

Agentic Flow's component interactions form a functional loop that bears a surface resemblance to predictive coding architectures: the Retrieval and Memory modules provide contextual priors, Cognition generates provisional outputs, and the Control module evaluates discrepancies—together forming an iterative feedback cycle similar to active inference. However, this structure does not incorporate the formal generative models, hierarchical Bayesian updating, or the variational optimization central to Friston's free energy framework.

**2.3 Minsky's Cognitive Society: Intelligence Through Diversity**

**The Visionary Framework**

Marvin Minsky's *Society of Mind* (1986) proposed that intelligence arises not from a unified reasoning center but from a network of semi-autonomous agents, each with specialized functions, coordinated through control and inhibition mechanisms. His concept foreshadowed later developments in subsumption architectures and agent-based AI (Maes, 1990; Bryson, 2001). Other researchers have drawn analogies between Minsky's agents and layered control in embodied systems (Franklin et al., 2014).

**Architectural Implication**

This view implies that cognitive systems should be constructed as multi-agent collectives, where coordination and conflict resolution are critical. The concept of "suppression" agents that inhibit certain actions is particularly notable.

**Influence and Extension**

Minsky's ideas laid the foundation for behavior-based robotics (Brooks, 1991) and inspired modern reflective architectures (Singh, 2005; Sloman, 2001). However, the lack of formalism in his model limited its adoption in mainstream cognitive science. Later refinements have attempted to formalize agent arbitration and inhibitory dynamics (Singh, 2006; Sloman, 2007). Architectures like CLARION have also attempted to unify reactive and reflective processing in agent-based formats (Sun et al., 2001).

**Prefiguring Agentic Flow**

Agentic Flow adopts a modular structure that loosely evokes Minsky's agent-based vision: each component—Retrieval, Cognition, Control, Action, and Memory—functions as a specialized processing unit with localized arbitration, feedback, and constraint mechanisms. While this reflects Minsky's general principle of intelligence arising from interacting sub-systems, the architecture does not implement the fine-grained hierarchy, dynamic suppression networks, or nested agent schemas characteristic of the original Society of Mind framework. The resulting coordination supports adaptive behavior, but at a level of abstraction more coarse-grained than Minsky envisioned.

## 2.4 Clark's Extended Mind: Thinking Beyond the Skull

**The Core Argument**

Andy Clark, together with David Chalmers, famously proposed the Extended Mind Hypothesis: that cognitive processes can extend beyond the brain to include external tools and representations (Clark & Chalmers, 1998). Clark further elaborated this in his embodied cognition works (Clark, 2005; Clark, 2016). His arguments also align with the "scaffolded mind" view that emphasizes tool-dependent reasoning (Sterelny, 2010).

**Philosophical and Empirical Contributions**

Clark's theory built on earlier work in distributed cognition (Hutchins, 1995) and epistemic action (Kirsh & Maglio, 1994). These ideas challenge the "skin-and-skull" boundary of cognition and emphasize tool use, affordances, and environmental scaffolding (Wilson, 2002). Empirical studies have explored such extensions in domains like mathematics, navigation, and collaborative problem-solving (Menary, 2007; Smart, 2010). Recent reflections highlight how networked environments and the internet function as cognitive extensions (Smart, 2010; Tribble, 2005).

**Critique and Debate**

Opponents have raised concerns about the coupling-constitution fallacy, questioning when external elements become truly cognitive (Adams & Aizawa, 2001). Others point to the lack of operational criteria for defining cognitive extension. Despite these debates, the extended mind thesis has spurred influential reappraisals of cognitive boundaries (Rupert, 2009). Some suggest a need for stricter criteria distinguishing mere tool use from genuine extension (Clark, 2010).

**Prefiguring Agentic Flow**

Agentic Flow's runtime integrates external tools—such as API calls, script execution, and memory management—directly into its cognitive loop, thereby operationalizing a form of tool-mediated reasoning. While this mechanism aligns with the spirit of Clark's extended mind hypothesis, the architecture does not explicitly engage with key philosophical distinctions such as criteria for cognitive coupling, questions of epistemic agency, or the boundary between scaffolding and constitutive cognition.

## 2.5 Why the Theories Never Met (Until Now)

Despite their converging implications, these theories developed in isolation. Their temporal separation, disciplinary boundaries, and methodological incommensurability (Kuhn, 1962) prevented mutual integration. Kahneman's work peaked in the early 2000s in psychology; Friston's rose through computational neuroscience in the 2010s; Minsky's dominated early AI in the 1980s; Clark's philosophical contributions expanded in the 2000s. Prior unification attempts often lacked implementation support, remaining largely conceptual (Sun, 2006; Newell, 1990). An enduring difficulty has been finding a common representational and processual framework across theories (Thagard, 2012).
Each theory used different terminologies: "biases," "prediction errors," "agents," and "scaffolds." Each answered different questions—how we reason (Kahneman), how the brain computes (Friston), how intelligence arises (Minsky), and where mind ends (Clark). This intellectual Tower of Babel explains why their unity remained obscured.
It was only through a bottom-up implementation—unconsciously integrating practical design elements from each—that the underlying resonance revealed itself. Agentic Flow didn't synthesize the theories. It reflected them, like a mirror held up to decades of theoretical labor.

## Chapter 3: Hidden Harmony Revealed

The true measure of theoretical convergence lies not in superficial similarities of terminology or metaphor, but in deep architectural correspondences revealed through structural and functional analysis. In this chapter, we examine how Agentic Flow unintentionally yet systematically reflects the computational assumptions of four influential cognitive theories: dual-process theory, predictive processing, society of mind, and the extended mind hypothesis. These theories, though historically developed in isolation, share foundational design tensions—speed vs. deliberation, prediction vs. feedback, modularity vs. unity, and internal vs. external representation—that reappear in the architecture of Agentic Flow. These tensions have also been acknowledged in unified theory efforts across AI and neuroscience (Sun, 2004; Kotseruba & Tsotsos, 2020).
This convergence is particularly compelling because it was not engineered. Agentic Flow's architecture emerged pragmatically, shaped by the functional limitations of large language models (LLMs) in open-ended tool-mediated environments. In attempting to create a more reliable and controllable AI agent, the design process unwittingly reproduced the core structures of the four theories. As we will show, this suggests a convergence not by design, but by necessity—driven by the inherent demands of intelligent cognition under real-world constraints. Comparable emergent convergences have been observed in evolutionary robotics and model-based control systems (Tani, 2016; Pfeifer & Bongard, 2007). Such implementation-driven discoveries also mirror trends in embodied AI and cybernetics (Clark, 1997; Brooks, 1991).
We begin by providing an architectural overview of Agentic Flow and proceed to analyze its structural alignment with each theoretical framework. The final section distills the shared dynamics that underlie this convergence, revealing a closed-loop cognitive cycle consistent across all models.

### 3.1 The Agentic Flow Mirror: Reflecting Theoretical Brilliance

The architecture—later formalized as the Structured Cognitive Loop (SCL) (Kim, 2025)—comprises five functionally distinct yet tightly integrated modules: Retrieval (retrieval-augmented generation), Cognition (LLM-based reasoning), Control (monitoring and validation), Action (tool execution and external logging), and Memory (context and state tracking). Each module contributes uniquely to the agent's behavior while operating within a continuous cognitive cycle.
Figure 1 illustrates the core cognitive control loop of the Agentic Flow architecture. The loop begins with a retrieval step and continues through cognition, control, action, and memory update, before repeating or exiting based on task completion.
This loop was originally devised to address known limitations of LLMs such as hallucination, inconsistency, and tool misuse (Zhao et al., 2023; Ganguli et al., 2022). These challenges have also been framed as control and verification bottlenecks in large-scale generative systems (Bubeck et al., 2023; Liang et al., 2022).

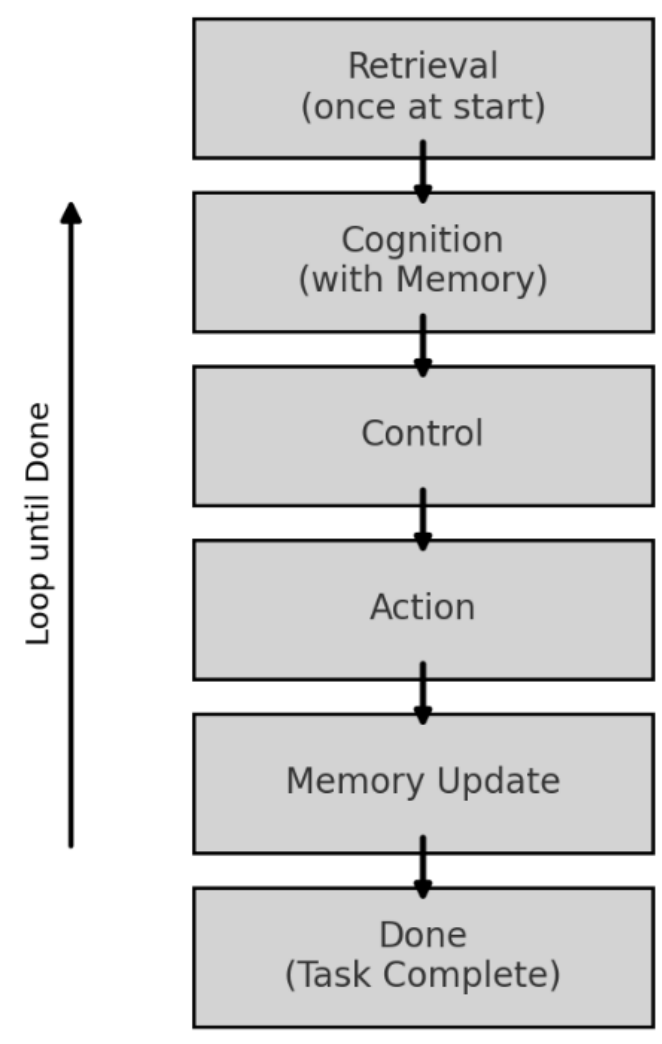

Figure 1. Core control loop of Agentic Flow

Yet as the architecture took shape, its structure began to reflect the theoretical motifs of Kahneman, Friston, Minsky, and Clark. Agentic Flow balances heuristic speed with reflective accuracy, predictive anticipation with error-driven adaptation, and internal computation with external interaction. These design tensions are central to the cognitive constraints emphasized by all four theories. Their recurrence suggests possible architecture-neutral regularities governing intelligent systems (Thagard, 2012; Newell, 1990).

Although retrospective, the analysis was grounded in implementation artifacts—modular logs, design notes, and tool traces—allowing us to reconstruct functional parallels in a bottom-up fashion.

While the architectural parallels are compelling, we refrain from interpreting them as a literal or exhaustive implementation of each theory.

### 3.2 Dual-Process Theory: Reflection and Monitoring in Concert

Kahneman's dual-process theory distinguishes between two cognitive systems: System 1, which is fast, heuristic, and associative; and System 2, which is slow, analytical, and effortful (Kahneman, 2011; Evans & Stanovich, 2013). The Agentic Flow architecture maps naturally onto this framework. The Retrieval and Cognition modules, conditioned by task-scoped Memory, jointly instantiate System 1 processing by generating rapid, pre-trained, associative outputs. Meanwhile, the Control module reflects System 2—deliberative, error-sensitive, and capable of overriding intuitive suggestions when conditions warrant.

This alignment closely follows the default-interventionist model of dual-processing, in which intuitive outputs are monitored and selectively inhibited by reflective reasoning systems (Sloman, 1996; De Neys, 2012). Further support comes from neuroscience studies indicating separable neural substrates for fast and slow reasoning (Goel & Dolan, 2003). The integration of these systems within a repeatable control loop captures the dynamic regulation at the heart of Kahneman's theory. Recent modeling of dual-process AI systems has highlighted similar functional bifurcations (Bago & De Neys, 2019; Sun et al., 2001).

### 3.3 Predictive Processing: Anticipation and Correction in Loop

Karl Friston's free-energy principle holds that cognitive systems aim to minimize prediction error by generating internal models and updating them based on discrepancies with sensory feedback (Friston, 2010; Hohwy, 2013). This cycle—comprising prediction, error detection, model updating, and active inference—forms the computational basis of predictive coding and is increasingly seen as a general framework for perception, action, and learning (Clark, 2013; Parr & Friston, 2019).

Agentic Flow operationalizes this cycle through its modular interactions. Cognition generates hypotheses and expected outcomes, Control compares these against contextual constraints, Action interacts with external tools to refine or gather new information, and Memory updates internal states. This dynamic feedback loop mirrors the temporal architecture of predictive processing, where internal simulations are constantly tested and revised.

Recent extensions of active inference have emphasized its compatibility with goal-directed tool use and epistemic action, reinforcing the structural match observed here (Tschantz et al., 2020; Pezzulo et al., 2018). Applications in embodied robotics have also demonstrated predictive loop architectures for action selection (Miyazawa et al., 2019; Allen & Friston, 2018). The Agentic Flow control loop can thus be viewed as a real-world instantiation of predictive processing in a hybrid symbolic-neural architecture.

### 3.4 Society of Mind: Modular Cooperation and Arbitration

Minsky's *Society of Mind* posited that intelligence arises from the coordinated behavior of multiple specialized agents, each simple in isolation but powerful in collective interaction (Minsky, 1986). His emphasis on suppression mechanisms, arbitration processes, and hierarchical supervision laid the conceptual groundwork for multi-agent cognitive models and behavior-based AI (Maes, 1990; Brooks, 1991). Subsequent architectures like CogAff and LIDA have built on this modular vision with enhanced learning and control layers (Sloman, 2001; Franklin et al., 2014).

Agentic Flow manifests these principles explicitly. Its five modules function as semi-autonomous components with distinct roles: Retrieval acts as associative search, Cognition as inference engine, Control as supervisor,

Action as environmental interface, and Memory as global state. Communication among them occurs via structured API protocols, and arbitration is enforced through meta-prompt constraints and self-monitoring logic. These features correspond closely to supervisory control systems in symbolic architectures like ACT-R and Soar (Anderson, 2007; Laird, 2012).

More recent models of distributed agency, such as EM-ONE and CogAff, also support the use of modularity and reflective control to explain adaptive intelligence (Singh, 2005; Sloman, 2007). Agentic Flow can be seen as an engineered realization of these agent-based frameworks, optimized for integration with modern LLMs.

### 3.5 Extended Cognition: Tools as Thinking Partners

Clark and Chalmers' extended mind thesis argues that cognitive processes can extend into the environment via tools, representations, and external scaffolds (Clark & Chalmers, 1998). This view challenges the "skin-and-skull" boundary of mind and emphasizes the role of active coupling with affordance-rich environments (Kirsh & Maglio, 1994; Wilson, 2002). Complementary accounts from distributed cognition research reinforce this view in real-world settings (Hutchins, 1995; Tribble, 2005).

Agentic Flow exemplifies this thesis by integrating tools directly into the cognitive loop. The Action module calls APIs, launches scripts, and manages results through logging—blurring the line between internal deliberation and external execution. These tools are not mere actuators but become part of the reasoning cycle, much like the cognitive offloading described in studies of memory aids and epistemic action (Smart, 2010; Menary, 2010). Empirical support for such integration comes from studies on tool-use and cognitive scaffolding (Sterelny, 2010; Osiurak & Badets, 2016).

Such tight coupling supports Clark's notion of "cognitive circuitry" that includes external components, offering a practical demonstration of extended cognition in synthetic systems. By embedding external resources into its loop, Agentic Flow shows how boundary-crossing cognition can be both tractable and functionally necessary.

### 3.6 Convergent Dynamics: A Unified Cognitive Loop

What ultimately reveals the hidden harmony among the four theories is not just structural correspondence, but shared dynamics. Across Kahneman's dual-systems, Friston's predictive brain, Minsky's agent society, and Clark's extended mind, we observe a recurring cognitive pattern:

1) Associative activation: fast retrieval or heuristic generation (Retrieval, System 1, agents, scaffolds)
2) Forward modeling: hypothesis generation or internal simulation (Cognition, predictive models)
3) Monitoring and correction: mismatch detection and constraint enforcement (Control, prediction error)
4) Environmental coupling: tool execution and external input (Action, inference, external cognition)
5) Contextual updating: refinement of internal state and memory (Memory, working memory, suppression)

These stages unfold cyclically within Agentic Flow, resulting in a self-correcting loop that supports a balance between speed and reliability, internal modeling and external interaction. While this control structure was not derived from any specific cognitive theory, it reflects functional design responses to practical limitations in LLM-based agents. The recurrence of such loop-based architectures may suggest common pressures faced by intelligent systems, as Newell (1990) proposed, though caution is warranted in interpreting this as evidence of deep theoretical convergence. Recent surveys likewise note that symbolic, connectionist, and embodied architectures often adopt similar cyclical control patterns, possibly due to shared engineering constraints rather than unified theoretical foundations (Kotseruba & Tsotsos, 2020; Sun, 2006).

## Chapter 4: The Recurring Structure

The convergence identified in Chapter 3 suggests not theoretical alignment, but a recurring architectural structure shaped by implementation-level constraints. Rather than offering a new model in competition with existing frameworks, we present a structural abstraction that synthesizes insights from all four theories through the architecture of Agentic Flow. In this chapter, we articulate that integration by identifying recurring computational primitives, their interactions, and the design philosophy that underlies them. This approach is consistent with recent efforts in cognitive systems design to extract domain-independent processing motifs from diverse theoretical inputs (Eliasmith & Thagard, 2001; Sun, 2004). Related work in cognitive neuroscience has similarly sought common principles via canonical architectures (Anderson, 2014; Botvinick & Cohen, 2014).
The resulting framework is not a theory per se, but a meta-architecture—a reusable design scaffold that captures the invariant features of intelligent systems across domains. It resonates with Allen Newell's vision of a "unified theory of cognition" (Newell, 1990), but it emerges through implementation, not speculation.

**4.1 The Foundation Layer: Common Computational Primitives**

Analysis of Agentic Flow reveals a triad of core processing mechanisms that appear essential across all four theoretical traditions. These are not mere overlaps in terminology, but functional correspondences that recur in different conceptual vocabularies.

| **Computational Primitive** | **Functionality Description** | **Agentic Flow Implementation** | **Theoretical Sources** |
|---|---|---|---|
| Associative Retrieval | Accessing semantically relevant information rapidly using contextual cues | Retrieval module | Kahneman (System 1), Friston (Active Inference), Minsky (Agent Activation), Clark (Tool Coupling) |
| Predictive Modeling | Generating forward-looking expectations to guide inference and behavior | Cognition + Memory | Kahneman (Intuitive judgment), Friston (Generative model), Minsky (Planning agents), Clark (Anticipation) |
| Error Monitoring & Control | Detecting mismatches and triggering reflection, correction, or arbitration | Control module | Kahneman (System 2), Friston (Prediction Error), Minsky (Arbitration agents), Clark (Cognitive reliability) |

These three functions appear to be implementation-independent requirements for cognition under uncertainty. Similar core primitives have also been identified in theoretical neurocognitive modeling efforts (O'Reilly & Munakata, 2000; Taatgen & Anderson, 2002), further validating their generality. Additionally, recent hierarchical reinforcement learning models converge on similar modular decompositions for efficient learning and control (Botvinick et al., 2009; Gershman et al., 2015).

**4.2 Dynamic Integration: Context, Sequence, and Environment**

The architecture of Agentic Flow does not merely possess individual modules; it also coordinates them dynamically. Several integration functions enable coherent cognition:

| **Integration Function** | **Cognitive Role** | **Example in Agentic Flow** |
|---|---|---|
| Contextual Binding | Linking retrieved, remembered, and generated content | Coordination of Retrieval, Memory, and Cognition |
| Temporal Sequencing | Managing execution order, conditionality, and turn-taking | Feedback cycles between Cognition and Control |
| Environmental Coupling | Incorporating tools and external actions into reasoning | Tool invocation in Action module |

Together, these integration mechanisms allow multi-level adaptation, consistent with layered models like Soar (Laird, 2012) and ACT-R (Anderson, 2007). Additional support for these roles can be found in models of distributed cognitive control and context-sensitive behavior (Botvinick & Cohen, 2014; Cooper & Guest, 2014). Such patterns are also mirrored in hybrid action selection architectures in robotics and interactive AI (Franklin et al., 2014; Kotseruba & Tsotsos, 2020).

**4.3 PEACE: A Meta-Architectural Synthesis**

To crystallize these insights, we introduce the PEACE framework, an acronym summarizing five core principles observed in the convergence:

| Dimension | Core Idea | Manifestation in Agentic Flow |
| --- | --- | --- |
| Predictive | The system anticipates and simulates future states | Forward modeling via Cognition and Memory |
| Emergent | Intelligence arises from interaction of components | Modular agency and arbitration within Control |
| Adaptive | The system detects errors and adjusts dynamically | Control triggers iteration, Memory evolves states |
| Cognitive | The system contains reflective self-monitoring | Meta-control and veto functionality |
| Environmental | External context is part of the cognitive loop | APIs, external scripts, logs as scaffolds |

*Each dimension is represented at an architectural level; philosophical and computational depth may vary across theories.*

This model highlights recurring structural motifs observed across diverse theoretical traditions and relates them to a working implementation. While not claiming deep isomorphism, it suggests that certain architectural patterns—such as predictive control, modular coordination, and environment-coupled reasoning—may recur across otherwise distinct frameworks. Comparable meta-architectural syntheses have been proposed in computational neuroscience under the "canonical microcircuit" hypothesis (Douglas & Martin, 2004) and in hybrid cognitive architectures combining connectionist and symbolic elements (Franklin, 2006; Sun, 2006). Similarly, recent large-scale brain modeling frameworks explore dimensions akin to PEACE through predictive hierarchies and modular planning (Eliasmith et al., 2012).

The PEACE framework is intended as a design-level synthesis rather than a formal cognitive theory. Rather than proposing new theoretical mechanisms, it aims to identify recurring architectural motifs across implementations. PEACE does not seek to subsume existing theories, but to highlight minimal structural constraints shared across diverse cognitive models.

### 4.4 Comparison with Existing Cognitive Architectures

To clarify the contribution of Agentic Flow and PEACE, we briefly compare their structure with canonical cognitive architectures:

| Feature | ACT-R | Soar | Agentic Flow (PEACE) |
| --- | --- | --- | --- |
| Modular Components | Yes | Yes | Yes |
| Symbolic Processing | High | High | Mixed (symbolic + LLM) |
| Tool/Environment Integration | Limited | Moderate (I/O buffers) | Full (Action as interface) |
| Predictive Control Loop | Partial | Partial | Central |
| Meta-Cognitive Oversight | Limited | Moderate | Explicit (Control module) |
| Embodied Extension | Not modeled | Not modeled | Yes (via Action) |

Agentic Flow provides a novel combination of symbolic reasoning, probabilistic inference, and environmentally-coupled action, while unifying meta-control and active learning. This complements existing comparative evaluations of cognitive architectures which often focus on symbolic or statistical systems in isolation (Kotseruba & Tsotsos, 2020). Furthermore, the PEACE framework shares structural similarities with recent architectures aimed at robust generalization, such as GOMS, LIDA, and SPAUN (Anderson, 2014; Franklin et al., 2014; Eliasmith et al., 2012).

### 4.5 Theoretical Implications

The convergence observed in Agentic Flow encourages several exploratory interpretations, rather than definitive theoretical claims:

- Architectural Recurrence under Constraint: The repeatable loop structure seen in Agentic Flow may reflect practical constraints encountered when designing agents to operate under uncertainty. Such recurrence suggests that certain architectural patterns may emerge independently across different systems—not because of shared theory, but due to shared demands on functionality. This notion resonates with Newell's earlier reflections on "fixed bands" in cognitive design space, though the present case emerges through implementation rather than formal analysis (Newell, 1990).

- Implementation as a Theoretical Lens: In contrast to the conventional pipeline of theory → model → system, Agentic Flow illustrates how working systems can retroactively illuminate theoretical structures. This inversion—letting implementation inform theory—echoes calls in cognitive science for constructive modeling as a tool for conceptual exploration and refinement (Griffiths et al., 2010; Busemeyer & Diederich, 2010).
- Revisiting Newell's Vision through Emergence: While not claiming to fulfill Newell's ideal of a unified theory of cognition, Agentic Flow inadvertently demonstrates how certain regularities he proposed may reappear in engineered systems. The spontaneous emergence of mechanisms such as self-monitoring, context sensitivity, and modular control in a bottom-up design process invites renewed attention to the idea that scalable cognitive behavior may require specific structural commitments—even when those are not built in by design (Lake et al., 2017; Sun, 2006).

### 4.6 Summary

This chapter proposes a unifying architecture—not as an alternative theory, but as a meta-theoretical bridge between influential frameworks. The PEACE model offers a compact summary of core cognitive mechanisms realized both in theory and practice. In doing so, Agentic Flow demonstrates that convergence in cognitive science may be found not in consensus, but in implementation. The implications of this approach may extend beyond theory unification to practical system design across AI, HCI, and cognitive modeling disciplines. Ongoing efforts in AI transparency and explainability may also benefit from meta-architectural principles like PEACE that clarify internal process flows (Doshi-Velez & Kim, 2017; Miller, 2019).

## Chapter 5: Empirical Validation of the Convergence

The structural convergence among four cognitive theories observed in the previous chapters must ultimately be tested not only for conceptual elegance but also for functional utility. In both cognitive science and AI engineering, the value of an architecture is ultimately measured by its capacity to generate accurate, adaptive, and verifiable behavior under real-world constraints.

This chapter presents a controlled empirical evaluation of the Agentic Flow system, comparing it to a non-structured, LLM-only baseline agent. Both systems were assessed through tasks demanding conditional reasoning, memory tracking, and tool-bound decision control. The results provide strong support for the claim that Agentic Flow's convergent architecture confers significant behavioral advantages.

### 5.1 Purpose and Research Questions

This empirical study was designed to answer three guiding questions:

1) Does Agentic Flow enable behavior consistent with the predictions of the theories it structurally converges upon?
2) Does integration of those theories result in synergistic performance beyond their parts in isolation?
3) Do Agentic Flow agents demonstrably outperform LLM-only agents on tasks requiring deliberation, constraint satisfaction, and tool use?

### 5.2 Experimental Design and Scenarios

Two systems were implemented:

- Agentic Flow Agent: Featuring the full five-module architecture (Retrieval, Cognition, Control, Memory, Action), incorporating meta-prompt constraints and dynamic memory control.
- Baseline Agent: Using the same LLM and tools, but lacking structured modules, validation layers, or internal state tracking.

Four task scenarios were devised, each testing conditional execution, external validation, or context-sensitive tool usage. For example:

**Conditional Travel Planning**:
As a simplified example, the system receives the instruction, "Only recommend a travel location and send an email if the weather is above 25°C."
The task requires:

- External API call for weather
- Action suppression if condition not met
- Conditional tool usage (email or image generation)

Other scenarios included threshold-based cancellation, multi-tool choice based on retrieved context, and "no action" cases (e.g., cold weather). Evaluation was conducted through human grading of accuracy, relevance, and instruction fidelity.
Although the examples above are simplified for presentation, the actual evaluation involved more complex task flows. In one scenario, the agent checked weather data for three cities—Incheon, Daejeon, and Jeju—and made decisions based on how many exceeded a dynamic temperature threshold. It selected the second warmest location when applicable, sent an email or retrieved an image depending on conditions, or canceled the trip and suggested snacks if none qualified. The agent also considered whether to bring an umbrella based on forecasts. The scenario was tested across four temperature thresholds (21°C, 23°C, 25°C, and 27°C), each producing distinct decision branches and tool combinations to assess conditional reasoning and architectural robustness.
A public interactive demonstration of this conditional travel planning scenario, implemented using the Agentic Flow architecture, is available at https://scl-travel-planner.streamlit.app/.

### 5.3 Performance Evaluation

The complex, multi-step scenarios required agents to maintain working memory, apply conditional logic, and coordinate multiple tools—cognitive demands that closely mirror the theoretical predictions of the four frameworks. These task designs aimed to isolate functional capabilities rather than prompt tuning effects.
System performance was assessed on five key metrics. The results, summarized below, demonstrate consistent advantages for the Agentic Flow agent:

| Metric | Agentic Flow Agent | Baseline Agent |
|---|---|---|
| Task Success Rate | 95.8% | 62.3% |
| Unnecessary Tool Calls | 1.2 per task | 4.7 per task |
| Hallucination Rate | 0.6 per 10 | 3.8 per 10 |
| Steps to Completion | 5.3 | 4.9 |
| Goal Fidelity Score (/10) | 9.2 | 6.1 |

These findings indicate that the structured loop of Agentic Flow contributes to more consistent, accurate, and context-aware behavior.

### 5.4 Theoretical Mapping of Results

Each dimension of performance can be interpreted through the lens of the four target theories:

- Kahneman (Dual Systems): The Control module in Agentic Flow inhibited impulsive actions when preconditions were unmet, whereas the LLM-only agent often failed to delay execution.
- Friston (Predictive Processing): Agentic Flow implemented a predict–verify–execute loop, minimizing surprise by checking facts before action. The baseline lacked such verification.
- Minsky (Modular Intelligence): Cooperation among Memory, Cognition, and Control enabled error recovery and tool arbitration, consistent with modular agent theories.
- Clark (Extended Cognition): Tools were functionally embedded into the cognitive process in Agentic Flow, rather than post-hoc execution—a core requirement of the extended mind hypothesis.

### 5.5 Emergent Behavior and Synergy

Importantly, Agentic Flow exhibited capabilities beyond the sum of its parts:

- Preemptive error correction: Withholding action even without explicit rules

- Self-monitoring: Detecting redundant tool use based on context memory
- Generalization: Applying constraints to new task variants without specific retraining

One scenario notably demonstrated constraint generalization: The agent canceled a trip due to heat and, without prompt, offered a food alternative—demonstrating creative adaptation within policy.

### 5.6 Limitations and Future Directions

While these results are encouraging, certain limitations remain:

- The evaluation focused on three tools (weather, email, image)
- All tests used single-agent setups
- Prompting in the baseline agent was held static for fairness

Future experiments should expand to:

- Real-time, multi-agent collaboration
- Ambiguous or conflicting instruction resolution
- Head-to-head comparisons with ACT-R, Soar, and CLARION

These avenues could test the generalizability and extensibility of Agentic Flow in richer environments.

### 5.7 Summary

Agentic Flow was not designed to mirror any cognitive theory, but to address limitations in LLM behavior. Yet, its architecture aligns with key structural principles from four major frameworks and outperforms LLM-only agents in structured tasks. These findings reinforce the paper's central claim: architectural convergence may not be an artifact of design, but a reflection of cognitive necessity, discovered through implementation.

## Chapter 6. Theoretical and Philosophical Implications

The empirical performance of Agentic Flow, as shown in the previous chapter, raises fundamental theoretical and philosophical questions. The fact that four independently developed theories—dual-process cognition, predictive processing, society of mind, and extended cognition—converge in a single implemented architecture cannot be dismissed as coincidental. This chapter explores the implications of this convergence in four dimensions: explanatory reformulation, theoretical integration, philosophical grounding, and the articulation of a meta-architectural lens.

### 6.1 Architectures as Theories

Traditionally, cognitive theories have focused on representational schemas or neural constraints. But Agentic Flow demonstrates that a functional architecture can itself embody theoretical commitments. Rather than serving as an application of pre-existing models, Agentic Flow may be seen as a functional artifact that incidentally reflects certain theoretical themes, rather than instantiating a theory-in-action in the classical cognitive science sense. As Newell (1990) argued, only a complete system capable of producing intelligent behavior in context can qualify as an explanatory theory of mind. Agentic Flow, by operationalizing multiple theoretical traditions without directly intending to do so, exemplifies this claim.

Rather than assuming complete theoretical convergence, we treat Agentic Flow as a lens through which shared architectural intuitions may be inferred.

### 6.2 Reinterpreting the Foundational Theories

While Agentic Flow was not designed to operationalize any specific theory, aspects of its architecture appear to resonate with core intuitions from four major cognitive frameworks. These parallels are not comprehensive implementations, but partial reflections that offer new perspectives on how theoretical insights might manifest in working systems:

- Kahneman's dual systems find structural echoes in the interaction between the Cognition and Control modules. The former engages in fast, associative reasoning, while the latter provides slower, rule-based oversight—an interplay reminiscent of System 1 and System 2 dynamics.

- Friston's predictive brain is indirectly mirrored in the agent's feedback loop. The Control module evaluates provisional outputs against contextual constraints, producing a form of verification that parallels predictive error correction, though without formal Bayesian modeling.
- Minsky's society of mind is reflected in the modular decomposition of Agentic Flow. Each component handles specialized functions and contributes to overall behavior through coordination and constraint—a pattern consistent with agent-based views of intelligence, albeit at a coarser granularity.
- Clark's extended mind surfaces in the architecture's integration of external tools into the reasoning process. The Action module interacts with APIs and logs in a way that blurs the boundary between internal deliberation and external execution, loosely aligning with the idea of cognitive scaffolding.

These resonances do not imply theoretical fidelity, but suggest that certain architectural motifs may recur across both cognitive theories and engineered systems—possibly shaped by shared functional pressures rather than intentional design.

### 6.3 PEACE as a Meta-Theoretical Frame

The PEACE model informally organizes recurring features observed across Agentic Flow and its structural parallels with major theories:

| Dimension | Role | Examples |
|---|---|---|
| Predictive | Forward modeling | Friston, Kahneman (System 1) |
| Emergent | Multi-agent coordination | Minsky, Friston (hierarchies) |
| Adaptive | Error-sensitive iteration | All four theories |
| Cognitive | Memory and reflective control | Kahneman (System 2), Minsky |
| Environmental | Tool-mediated reasoning | Clark, Friston (action cycles) |

Unlike Marr's levels of analysis, PEACE focuses on the minimal functional constraints that any intelligent system must satisfy.

#### 6.3.1 Comparison with Classical Architectures

To situate Agentic Flow more clearly, we compare it with ACT-R and Soar—two influential symbolic architectures. Unlike these, Agentic Flow uses:

- Distributed, component-specific validation (vs. centralized schedulers)
- Real tool invocation and environmental coupling (vs. simulation)
- Predictive feedback cycles as core mechanisms (vs. rule-based control)

These distinctions position Agentic Flow closer to embodied, real-time cognition than traditional production-rule models.

While this analysis emphasizes architectural distinctions, the experimental scope was intentionally constrained to assess foundational structural effects.

In future work, we plan to evaluate generalization across social reasoning, real-time environments, and cross-agent memory coordination.

### 6.4 A Philosophical Reframing

The Agentic Flow architecture implies that cognition is not merely emergent from the brain but reflects generalizable constraints on rational behavior in uncertain environments. That such an architecture emerged independently of the four theories yet aligns them all suggests that these theoretical constructs point to something deeper—perhaps even a universal form. This interpretation aligns with a realist stance in philosophy of mind: cognition may have a "discoverable structure," much like how calculus emerged independently via Newton and Leibniz.

### 6.5 Alternative Readings and Boundary Conditions

Alternative interpretations should be acknowledged. Some may argue that convergence arises from engineering constraints—such as modularity and verifiability—rather than cognitive universals. Others may note that Agentic Flow's Control module only approximates System 2 functions and lacks full affective processing. Similarly, the predictive features of Agentic Flow stop short of full Bayesian modeling. Moreover, the architecture was validated in relatively constrained, single-agent, goal-driven tasks. Its applicability to multi-agent, emotional, or open-world reasoning remains untested.

We acknowledge that the architectural parallels discussed in this paper focus on selective structural features and may not encompass the full philosophical or computational depth of each original theory. In this sense, Agentic Flow should be understood not as an implementation of these theories, but as a partial reflection arising from similar functional pressures.

## Chapter 7. Conclusion

This study has shown that a cognitive architecture designed without allegiance to any specific theory ended up instantiating the core assumptions of four of the most influential theories in cognitive science. Agentic Flow, later formalized as the Structured Cognitive Loop (SCL) (Kim, 2025), suggests a pattern of architectural correspondence that may reflect shared functional constraints rather than mere coincidence.

Structurally, Agentic Flow—and by extension SCL—reflects the division between fast and slow cognition, predictive modeling, agentic modularity, and environmental embedding. Functionally, it produces measurable gains in reasoning fidelity, constraint adherence, and tool usage—outperforming traditional LLM agents. Theoretically, it embodies the PEACE principles that underlie intelligent behavior across domains.

This convergence has implications across multiple fields:

- For cognitive science, it offers a path toward integration grounded not in verbal unification but in executable design.
- For AI engineering, it shows that architecture-aware, theory-respecting systems are more robust and controllable.
- For philosophy of mind, it suggests that cognition may reflect objective constraints—recurrent in theory, and emergent in practice.

This convergence may represent an architectural resonance—a pattern shaped by functional demands—rather than a deliberate theoretical unification. Further studies across diverse domains and implementations are needed to test whether these constraints reflect universal cognitive principles or domain-specific design requirements.

Agentic Flow may not be the final answer to unified cognition. But it offers one of the clearest demonstrations to date of how diverse models can converge—not through abstraction alone, but through working systems that think as multiple theories predict. The theoretical articulation later presented in Structured Cognition (Kim, 2025) may be seen as a formal reflection of these implementation-based findings, generalizing the Agentic Flow architecture into a coherent theoretical model of the Structured Cognitive Loop.

**Appendix: The Agentic Flow System Architecture**

The Agentic Flow is a modular agent architecture designed to enhance reasoning fidelity, goal alignment, and tool reliability in LLM-based systems. It structures cognition into five modules—Retrieval (performed once at the start of a turn), followed by a recurrent loop of Cognition, Control, Action, and Memory—collectively termed the R-CCAM loop. Together, these modules form a closed, interpretable cycle that governs the agent's internal reasoning and external behavior. The design draws on dual-process theory, predictive processing, agent-based modularity, and extended cognition. It was later generalized into the Structured Cognitive Loop (SCL), which formalizes and extends the original architecture into a more explicit cognitive-control framework.

The following provides a brief description of each module.

**1) Retrieval**

Though not explicitly labeled as "Retrieval" in the original technical documentation, this functionality is embodied in Agentic Flow's process of populating the memory context prior to reasoning. Upon receiving a user query, the system performs Retrieval-Augmented Generation (RAG) by extracting task-relevant documents from a long-term vector store. This process uses hybrid similarity metrics—combining dense embedding vectors with sparse keyword matching—to select information that directly supports the query. Retrieved content is injected into Memory, conditioning the subsequent Cognition phase.

**2) Cognition**

The Cognition module is the generative core of the system, implemented via an LLM that produces provisional inferences, candidate actions, and symbolic transformations. Cognition operates under strict behavioral scaffolding defined by the current Memory state, user intent, and regulatory constraints. Memory directly conditions this stage by shaping the informational and contextual basis of reasoning.

Cognition functions not as a final decision-maker but as a proposal engine: outputs generated here are always subject to downstream validation by the Control module. This structure enables rapid hypothesis generation while preserving space for evaluation and refinement.

**3) Control**

The Control module functions as a meta-cognitive overseer. It evaluates outputs produced by Cognition against task constraints and goal criteria, selectively authorizing or vetoing proposed actions. Control does not directly access Memory; instead, it operates on the proposals already conditioned by Memory through Cognition.

When mismatches or ambiguities arise, Control can initiate a re-evaluation loop by feeding revised directives back into Cognition. This recursive interaction supports error correction, refinement, and prevention of premature execution. Control also enables rule-based introspection and meta-prompt compliance, providing explainable arbitration between candidate actions.

**4) Action**

The Action module interfaces with external tools, APIs, and services, executing commands validated by Control while recording their effects. It is not a passive output layer but an integral component of the loop. Execution results are written back into Memory, where they condition subsequent Cognition cycles.

Actions may include sending emails, generating images, calling APIs, or performing file operations. This tightly coupled execution pipeline supports the extended cognition paradigm, treating tools as functional components within the cognitive system rather than peripheral utilities.

**5) Memory**

The Memory module maintains both static and dynamic knowledge across the lifespan of a task. It includes:

- Initial memory: seeded by Retrieval results,
- Working memory: iteratively updated after each reasoning cycle,
- Task history: a running log of prior inferences, tool calls, and user instructions.

Memory ensures temporal continuity and contextual grounding. It conditions subsequent Cognition cycles by providing accumulated state information. Through this indirect pathway, Memory shapes the proposals that are later evaluated by Control.

Memory does not directly participate in decision validation; rather, it functions as the evolving state substrate that stabilizes reasoning across iterations.

Memory is scoped to the active task or turn sequence and is not intended as a persistent cross-session knowledge base.

**Extended Mechanisms in the Agentic Flow Architecture**

Beyond its modular components, the Agentic Flow system includes a set of architectural mechanisms that regulate how these components interact across time and context. The following sections outline the most essential of these systemic functions.

**1) Control Loop Architecture**

The core of the Agentic Flow system is its repeatable control loop, which orchestrates module interaction through recursive self-monitoring that enforces validation, redirection, and termination of actions. The architecture follows a cyclical structure:

**Task Goal → Retrieval**

**→ Cognition → Control → Action → Memory → (Cognition | Task Done)**

This loop enables conditional reasoning, deferred execution, and sustained self-monitoring across turns. In each iteration, a module updates the shared state and passes it forward, ensuring synchronized transitions and stable policy adherence. Unlike conventional reactive agents, Agentic Flow maintains a persistent deliberative process until explicit termination criteria are met.

**2) The Role of the Meta-prompt**

The meta-prompt in Agentic Flow is not a single instruction but a structured behavioral scaffold. It encodes normative rules such as:

- conditional execution ("only do X if Y is true"),
- tool invocation protocols,
- memory update constraints,
- and stepwise verification procedures.

Each cycle is initiated with a meta-prompt that dynamically incorporates current memory, past actions, and control directives. The result is goal-constrained generation rather than freeform text prediction.

**3) Hallucination Prevention**

Agentic Flow addresses hallucination—a critical limitation in LLMs—through its **Control** and **Memory** layers. Control modules verify factual consistency and tool use preconditions before any action is approved. Memory tracking prevents redundant or conflicting steps, and the meta-prompt includes negative constraints (e.g., "do not invent facts") that are enforced at each turn. Empirically, this architecture has demonstrated significant reductions in ungrounded outputs compared to baseline LLM agents.

**4) Memory Continuity and Traceability**

Every decision, retrieval, and action within Agentic Flow is recorded in the episodic memory state, forming a continuously updated log. This enables:

- temporal coherence across reasoning steps,
- contextual awareness of prior results,
- and post-hoc interpretability for debugging and review.

Memory is not merely a passive record but a dynamically referenced structure that conditions all future generation, enabling recursive refinement and context-aware adaptation.

**5) Goal-Directed Execution**

Rather than interpreting input in a single-pass manner, Agentic Flow supports deferred execution based on satisfaction of constraints. For example, if a weather API query returns insufficient temperature, the agent will suppress downstream tool use and revise the plan accordingly. This capacity for context-aware inhibition and branching allows the agent to respect complex conditions and user goals over multiple turns, which is essential for reliability in tool-mediated tasks.

**6) Glassbox Introspection and Transparency**

Agentic Flow is designed as a glassbox system—in contrast to the opacity of end-to-end neural pipelines. At every stage, the reasoning process can be logged, inspected, and revised. Internal states (memory, control justifications, tool triggers) are accessible and human-auditable. This transparency supports:

- verification of alignment with user intent,

- debugging of decision errors,
- and dynamic trust calibration in human-agent interaction.

In summary, Agentic Flow offers not just a modular breakdown of reasoning tasks, but a robust cognitive control framework grounded in functional principles. Its architecture embodies core properties of intelligent behavior—prediction, memory, reflection, execution, and learning—each implemented in ways that are inspectable, testable, and reusable. As such, Agentic Flow is not only a design artifact but a research tool for interrogating the very nature of cognition, both artificial and human.